%% file: main.tex
\documentclass[sigconf]{acmart}
    \pdfoutput=1
    \renewcommand\footnotetextcopyrightpermission[1]{} 
    \usepackage{bm}
    \usepackage{url}
    \usepackage{amsmath,epsfig}
    \usepackage{subfigure}
    \usepackage{algorithm}
    \usepackage{algorithmic}
    \usepackage{booktabs}
    \usepackage{multirow}
    \usepackage{epstopdf}
    \usepackage{graphicx} 
    \renewcommand{\vec}[1]{\mathbf{#1}}

\begin{document}
    \title{RGCNN: Regularized Graph CNN for Point Cloud Segmentation}
    
    \author{Gusi Te}
    \affiliation{
    \institution{Peking University}
    }
    \email{tegusi@pku.edu.cn}

    \author{Wei Hu}
    \affiliation{
    \institution{Peking University}
    }
    \email{forhuwei@pku.edu.cn}

    \author{Zongming Guo}
    \affiliation{
    \institution{Peking University}
    }
    \email{guozongming@pku.edu.cn}
    
    \author{Amin Zheng}
    \affiliation{
    \institution{MTlab, Meitu Inc.}
    }
    \email{zam@meitu.com}

    \begin{abstract}
        Point cloud, an efficient 3D object representation, has become popular with the development of depth sensing and 3D laser scanning techniques. It has attracted attention in various applications such as 3D tele-presence, navigation for unmanned vehicles and heritage reconstruction. The understanding of point clouds, such as point cloud segmentation, is crucial in exploiting the informative value of point clouds for such applications. Due to the irregularity of the data format, previous deep learning works often convert point clouds to regular 3D voxel grids or collections of images before feeding them into neural networks, which leads to voluminous data and quantization artifacts. In this paper, we instead propose a regularized graph convolutional neural network (RGCNN) that directly consumes point clouds. Leveraging on spectral graph theory, we treat features of points in a point cloud as signals on graph, and define the convolution over graph by Chebyshev polynomial approximation. In particular, we update the graph Laplacian matrix that describes the connectivity of features in each layer according to the corresponding learned features, which adaptively captures the structure of dynamic graphs. Further, we deploy a graph-signal smoothness prior in the loss function, thus regularizing the learning process. Experimental results on the ShapeNet part dataset show that the proposed approach significantly reduces the computational complexity while achieving competitive performance with the state of the art. Also, experiments show RGCNN is much more robust to both noise and point cloud density in comparison with other methods. We further apply RGCNN to point cloud classification and achieve competitive results on ModelNet40 dataset.     
    \end{abstract}

    

    \maketitle
    
    \section{Introduction}
    \label{sec:intro}
    \input{intro}

    \section{Related Work}
    \label{sec:related}
    \input{related}

    \section{Problem Statement}
    \label{sec:problem}
    \input{problem}

    \section{The Proposed RGCNN}
    \label{sec:method}
    \input{method}

    \section{The Proposed Loss Function and Theoretical Analysis}
    \label{sec:theory}
    \input{theory}

    \section{Experimental Results}
    \label{sec:results}
    \input{results}

    \section{Conclusion}
    \label{sec:conclude}
    \input{conclude}
    
    \newpage

    \bibliographystyle{plain}
    \bibliography{acmart}
    \end{document}

%% file: intro.tex
The development of depth sensors like Microsoft Kinect and 3D scanners 
like LiDAR has enabled convenient acquisition of 3D point clouds, a 
popular signal representation of arbitrarily-shaped objects in the 3D space.     
Point clouds consist of a set of points, each of which is composed of 
3D coordinates and possibly attributes such as color and normal. 
Thanks to the efficient representation, point clouds have been widely deployed in 
various fields, such as 3D immersive tele-presence, navigation 
for unmanned vehicles, free-viewpoint videos and heritage preservation \cite{tulvan16}. Hence, the analysis of point clouds, such as point cloud segmentation, becomes active research topics in order to exploit the informative value of point clouds.  

\begin{figure}[htbp]
    \centering
    \includegraphics[width=0.45\textwidth]{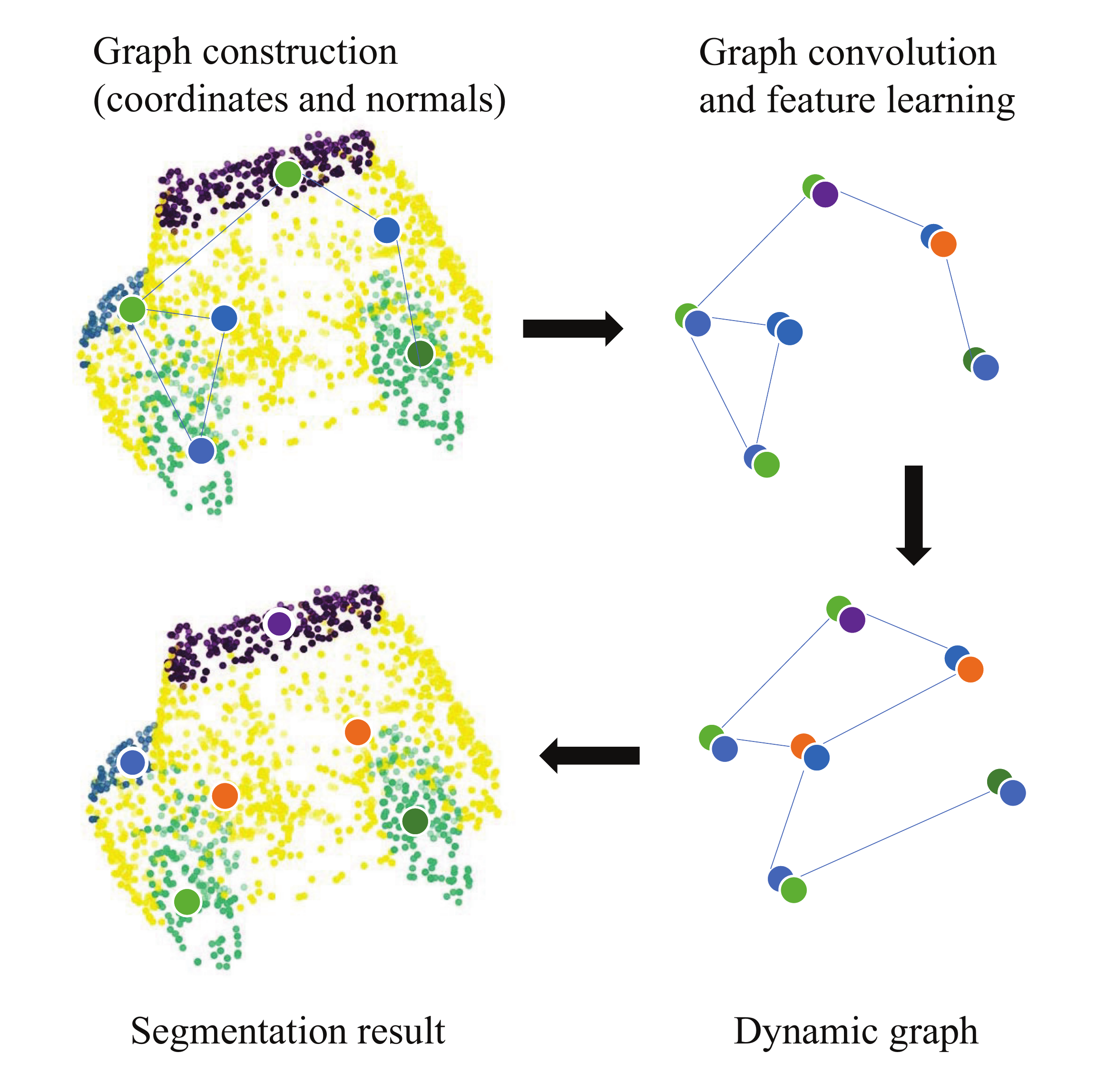}
    \vspace{-0.1in}
    \caption{Illustration of the RGCNN architecture, which directly consumes raw point clouds (\texttt{car} in this example) without voxelization or rendering. It constructs graphs based on the coordinates and normal of each point, performs graph convolution and feature learning, and adaptively updates graphs in the learning process, which provides an efficient and robust approach for 3D recognition tasks such as point cloud segmentation and classification. }
    \label{fig:title}
\end{figure}

Previous point cloud segmentation works can be classified into model-driven 
segmentation and data-driven segmentation. Model-driven methods include  
edge-based \cite{rabbani2006segmentation}, region-growing \cite{rusu2008towards} and model-fitting \cite{tarsha2007hough}, which are based on the prior knowledge of the geometry but sensitive to noise, uneven density and complicated structure. Data-driven segmentation, on the other hand, learns the semantics from data, such as deep learning methods \cite{qi2017pointnet}. Nevertheless, typical deep learning architectures require regular input data formats, such as images on regular 2D grids or voxels on 3D grids, in order to perform operations like convolution and pooling. For \textit{irregular} 3D point clouds, most previous works convert them to regular 3D voxel grids \cite{maturana2015voxnet} or collections of images \cite{su2015multi} before feeding them into typical convolutional neural networks (CNN). This, however, introduces quantization error in the conversion process and renders the resulting data unnecessarily voluminous. 


Recently, Graph Convolutional Neural Network (GCNN) has been proposed to generalize 
CNNs to graphs \cite{kipf2016semi}. The key idea is to consider the convolution of graphs in 
the spectral domain, leveraging on spectral graph theory \cite{chung97}. However, this requires the eigen-decomposition of graph Laplacian matrices \cite{chung97} that describe the connectivity of graphs, which is computationally expensive. 
Hence, several methods propose to approximate the convolution in the spectral domain by spatial filtering, such as Chebyshev polynomials \cite{hammond2011wavelets}, Lanczos method~\cite{susnjara2015accelerated}, Cayley polynomials \cite{levie2017cayleynets}, etc. Nevertheless, the graph Laplacian matrix is always \textit{fixed}, which is unable to represent the structures of dynamic graphs in the learning process. Also, though GCNN has shown its efficiency in semi-supervised classification, it hasn't been deployed to point cloud segmentation yet.   

In order to address the above problems, we propose a regularized graph convolutional neural network (RGCNN) for point cloud segmentation. As depicted in Fig.~\ref{fig:title}, RGCNN treats the features of points as graph signals, and takes the feature matrix and adjacency matrix of irregular point clouds as the input. Specifically, we choose the coordinates and normals of each point as the features to represent the underlying geometry of point clouds. The output is the per point segmentation labels for each point of the input. Leveraging on the basic framework of GCNN with truncated Chebyshev approximation, we design a three-layer GCNN with high-order Chebyshev polynomials. In particular, we incorporate a \textit{graph-signal smoothness prior} into the loss function, which regularizes the learning process. This essentially combines data-driven methods with model-driven ones. Further, we prove the \textit{spectral smoothing} property of this prior, which essentially enforces Laplacian smoothing in the spectral domain. Besides, instead of fixing the graph structure as in previous works (e.g., \cite{yi2016syncspeccnn}), we \textit{update} the graph Laplacian matrix in each layer, thus capturing the dynamic topology of graphs. We also prove the permutation-invariance property of the proposed RGCNN, so that when the input permutes, the output permutes in the same way. Finally, we extend the architecture of RGCNN for the application of point cloud classification. 
 
While details are presented in the paper, the key contributions are as follows:  
\begin{itemize}
\item To the best of our knowledge, we are the first to design GCNN for point cloud segmentation, which is suitable for consuming unordered 3D point clouds;
\item We regularize each layer of RGCNN by adding graph-signal smoothness prior in the loss function, and prove the spectral smoothing property of this prior; 
\item We update the graph Laplacian matrix in each layer of RGCNN, in order to adaptively capture the structure of dynamic graphs.
\item Extensive experiments show that RGCNN significantly reduces the computation complexity while achieving competitive results with the state of the art. It is also much more robust to both low density and noise in comparison with other methods.  
\end{itemize}

The rest of the paper is organized as follows. Section~\ref{sec:related} provides a review on previous works of point cloud coding and introduce GCNN as the basic framework. We present the problem statement of point cloud segmentation in Section~\ref{sec:problem}, and then elaborate on the proposed RGCNN in Section~\ref{sec:method}. The proposed loss function and theoretical analysis is discussed in Section~\ref{sec:theory}. Next, performance evaluation and comparison is presented in Section~\ref{sec:results}. Finally, we conclude the paper in Section~\ref{sec:conclude}.

%% file: related.tex
We will first review previous works on point cloud segmentation, and then introduce GCNN, which inspires the proposed method. 

\subsection{Point Cloud Segmentation}
Previous works on point cloud segmentation can be mainly classified into two categories: model-driven methods and data-driven methods. 

\textbf{Model-driven methods:}
This class of approaches segment point clouds by assuming certain models of the underlying geometry. According to different models, they are further categorized as follows.
\begin{itemize}
    \item \textbf{Edge-based methods}: Rabbani et al. propose an edge-based method in \cite{rabbani2006segmentation}, which outlines the borders of different regions by edge detection and then groups points inside the borders to deliver final segments. Jiang et al. \cite{jiang1996fast} propose scan-line grouping methods to represent surfaces, and divide them by edges. This achieves good performance on range images, but is unsuitable for point clouds with uneven density. Although edge-based methods are fast, they are sensitive to noise and uneven density.
    
    \item \textbf{Region growing methods}: Starting from one or more points with specific characteristics, these methods grow around adjacent and similar points. Further, these methods can be divided into top-down approaches and bottom-up approaches. The difference includes the initial choice of points and how they grow afterwards. The main disadvantage lies in the selection of initial points and in the presence of complicated structures.
    
    \item \textbf{Model fitting methods}: This category is based on the observation that many man-made objects can be decomposed into geometric primitives like planes, cylinder and spheres. Points that conform to some primitive shapes are treated as one segment. Two popular algorithms include the Hough Transform \cite{tarsha2007hough} and the Random Sample Consensus approach \cite{chen2014methodology}. However, details may fail to be modelled into easily recognizable geometric primitives.   
\end{itemize}

Model-driven methods are primarily limited by the assumed prior knowledge. Also, objects with complex structures are great challenges for small-scale model datasets.

\textbf{Data-driven methods}
This class is concerned with Artificial Intelligence algorithms based on empirical and training data. As 3D point clouds are irregular, traditional data-driven methods cannot be applied directly. Preprocessing, such as feature extraction, is thus required. Features of high quality can greatly improve the learning efficiency.

\begin{itemize}
    \item \textbf{Segmentation based on clustering}: Clustering groups (3D) points into clusters based on attributes or features. Biosca et al. deploy unsupervised clustering and fuzzy algorithms for laser point clouds \cite{biosca2008unsupervised}. Filin et al. proposes an approach using normal vectors derived from a neighborhood system called slope adaptive \cite{filin2002surface}, where the slopes of normal vectors and height difference between each point and its neighbors are applied as features. Ma et al. exploit spectral clustering for point cloud segmentation \cite{ma2010point}, and propose a novel approach to find k-nearest-neighbors for graph construction. Clustering-based segmentation achieves better performance than model-driven methods under complex scenes. However, there lacks the view of local information.
    
    \item \textbf{Deep learning segmentation}: Deep learning methods have shown great potential in 3D shape recognition, such as view-based learning \cite{su2015multi}, 
3D ShapeNets \cite{wu20153d}, VoxNet \cite{maturana2015voxnet} and VoxelNet \cite{zhou2017voxelnet}, which are mainly designed for point cloud classification. 

Regarding point cloud segmentation, Qi et al. come up with PointNet, a neural network which consumes point clouds directly \cite{qi2017pointnet}. The key breakthrough is the proposed symmetric function applied to the raw point cloud data. However, PointNet processes each point identically and independently. Hence, PointNet++ is proposed in \cite{qi2017pointnet++} by introducing hierarchical grouping, which achieves better performance. 
    
%

\end{itemize}

\subsection{Graph Convolutional Neural Network}
As CNN only deals with data defined on regular grids, it is extended to graphs for irregular data, which is referred to as GCNN. The key challenge is to define convolution over graphs, which is difficult due to the irregularity. 

\begin{itemize}
    \item \textbf{Spectral methods}: The convolution over graphs is defined in the spectral domain, which is the multiplication of the signal on graph with the eigenvector matrix of the graph Laplacian matrix \cite{hammond2011wavelets, henaff2015deep}. The computation complexity, however, is high due to the eigen-decomposition of the graph Laplacian matrix in order to get the eigenvector matrix. Hence, it is improved by \cite{defferrard2016convolutional} through fast localized convolutions, where Chebyshev expansion is deployed to approximate graph Fourier transform (GFT). Susnjara et al. introduce the Lancoz method for approximation  \cite{susnjara2015accelerated}.
    \item \textbf{Spatial methods}: In this method , many techniques are introduced to implement convolution directly on each node and its neighbors. Gori et al. introduce recurrent neural networks that operate on graphs in \cite{gori2005new}. Duvenaud et al. propose a convolution-like propagation to acculmulate local features \cite{duvenaud2015convolutional}. Bruna et al. deploy the multiscale clustering of graphs in convolution to implement multi-scale representation \cite{bruna2013spectral}. Furthermore, Niepert et al. define convolution on a sequence of nodes and perform normalization afterwards \cite{niepert2016learning}. Spatial methods provide strong localized filters, but it also means it is difficult to learn the global structure.
\end{itemize}

Spectral GCNN has shown its efficiency in semi-supervised classification \cite{kipf2016semi}, which outperforms many state-of-the-art methods significantly on citation networks and node classification. However, to the best of our knowledge, we are the first to extend it to point cloud segmentation.

%% file: problem.tex
We design a graph neural network that directly undertakes irregular 3D point clouds as the input for segmentation. 
We represent a point cloud as a set of 3D points $ \{ \mathbf{p}_i\}_{i=1}^n $, where   
$ \mathbf{p}_i \in \mathbb{R}^m $ is a vector denoting the $ i $-th point's feature, such as 
coordinates, color, normal, etc. In the proposed method, we adopt the coordinates $ (x_i,y_i,z_i) $  and normal $ (n_i^x,n_i^y,n_i^z) $ as the feature for the $ i $-th point, i.e., $ \mathbf{p}_i = (x_i,y_i,z_i,n_i^x,n_i^y,n_i^z)^T $, and thus $ m=6 $. 

In the proposed RGCNN, the input is a $ n \times m $ feature matrix $ \mathbf{P} $ and a $ n \times n $ adjacency matrix $ W $. Assuming we have $ k $ semantic labels, RGCNN outputs $ n \times k $ scores $ \mathbf{S} $ for each of the $ n $ points and each of the $ k $ labels.

%% file: method.tex
We first present the overall RGCNN architecture, and then elaborate on important modules including graph construction, graph convolution and feature learning respectively. 

\subsection{RGCNN Architecture}
\begin{figure*}[htbp]
	\centering
	\includegraphics[width=\textwidth]{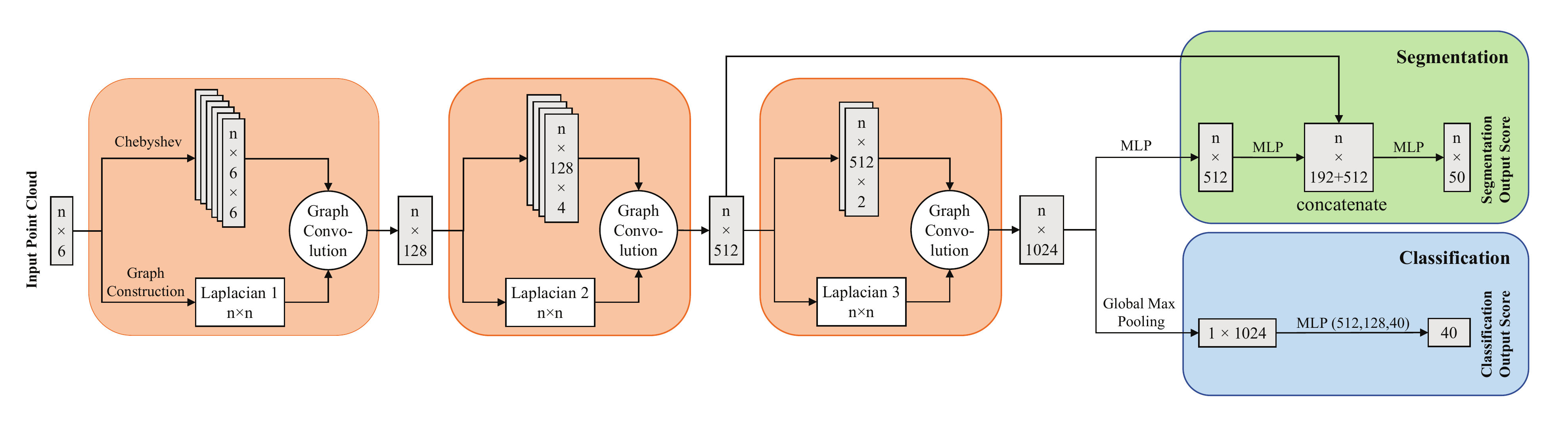}
	\caption{The architecture of the proposed RGCNN.}
	\label{fig:graph}
\end{figure*}
As depicted in Fig.~\ref{fig:graph}, RGCNN consists of one general model for extracting features and two branches for segmentation and classification tasks respectively. It takes raw point clouds with coordinates and normals as the input, learns local features by graph convolution and then outputs the segmentation or classification score. In the segmentation branch, we deploy graph convolution to aggregate features, and then concatenate features from different layers to represent both local and global features. The per-point label is finally given in the output layer. In the classification branch, we additionally deploy global max pooling to collect global features and use multilayer perceptron (MLP) to get the final score. Specifically, RGCNN has three regularized graph convolution layers. Each layer consists of graph construction, graph convolution and feature filtering, which are elaborated in order as follows. 

\subsection{Graph Construction}
As the graph construction has crucial effect on the efficiency of the network, we first discuss the proposed approach to construct graphs over point clouds. 

\textbf{Graph and Graph Laplacian.}~~~
We consider an undirected graph $ \mathcal{G}=\{\mathcal{V},\mathcal{E},\mathbf{A}\} $ composed of a vertex set $ \mathcal{V} $ of cardinality $|\mathcal{V}|=n$, an edge set $ \mathcal{E} $ connecting vertices, and a weighted \textit{adjacency matrix} $ \mathbf{A} $. $ \mathbf{A} $ is a real symmetric $ n \times n $ matrix, where $ a_{i,j} $ is the weight assigned to the edge $ (i,j) $ connecting vertices $ i $ and $ j $. We assume non-negative weights, \textit{i.e.}, $a_{i,j} \geq 0$. 

The Laplacian matrix is defined from the adjacency matrix. Among different variants of Laplacian matrices, the \textit{combinatorial graph Laplacian} used in \cite{Shen10,Hu12,Hu14} is defined as $ \mathcal{L}_c:=\mathbf{D}-\mathbf{A} $, where $ \mathbf{D} $ is the \textit{degree matrix}---a diagonal matrix with $ d_{i,i} = \sum_{j=1}^n a_{i,j} $. One \textit{normalized} graph Laplacian matrix is defined as $ \mathcal{L}=\mathbf{D}^{-\frac{1}{2}}\mathcal{L}_c\mathbf{D}^{-\frac{1}{2}} $, which is used in the sequel because of its normalization property.  

\textbf{Graph signal.}~~~
Graph signal refers to data residing on the vertices of a graph. In this paper, the graph signal is the features of each point in the point cloud, i.e., the feature vector  $ \mathbf{p}_i $ of the $ i $-th point.  

\textbf{Graph Construction.}~~~
Though there exist various ways to construct graphs, we choose complete graphs, which connect each point with all the other points in the point cloud and thus consider the relationship among all the points. The edge weight is defined based on the distance between features of points, which is able to measure the similarity among points in terms of structure. Specifically, the weight of an edge connecting points $ i $ and $ j $ is defined as 

\begin{equation}
a_{i,j}=
    \exp(-\beta\|\mathbf{p}_i-\mathbf{p}_j\|_2^2),
\end{equation}  
where $\beta$ is a scalar parameter. We empirically set $ \beta=1 $ in the experiments. 

\subsection{Graph Convolution}
The core of GCNN is graph convolution. Unlike images or videos, it is difficult to define convolution over graphs in the vertex domain, because a meaningful translation operator in the vertex domain is nontrivial to define due to the unordered vertices. Hence, inspired by \cite{defferrard2016convolutional}, we start from filtering of graph signals in the spectral domain, and then deploy Chebyshev approximation to reduce the computational complexity. 

\textbf{Spectral filtering of graph signals.}~~~
The convolution operator on a graph $ \ast_{\mathcal{G}} $ is first defined in the spectral domain \cite{bruna2013spectral}, specifically in the GFT domain. GFT is computed from the graph Laplacian matrix. As the graph Laplacian is symmetric and positive semi-definite, it admits a complete set of orthonormal eigenvectors. The GFT basis $\mathbf{U}$ is then the eigenvector set of the Laplacian matrix. The GFT of a graph signal $ \mathbf{x} $ is thus defined as $ \hat{\mathbf{x}} = {\mathbf{U} ^T}\mathbf{x} $, and the inverse GFT follows as $ \mathbf{x}' = \mathbf{U} \hat{\mathbf{x}} $.
 
Hence, the convolution between two graph signals $ \mathbf{x} $ and $ \mathbf{y} $ can be defined as the multiplication of the corresponding GFT coefficients, followed by the inverse GFT, i.e., 

\begin{equation}
\mathbf{x} \ast_{\mathcal{G}} \mathbf{y} 
= \mathbf{U}(\mathbf{U}^T\mathbf{x}) \odot (\mathbf{U}^T\mathbf{y}),
\end{equation}
where $ \odot $ is the element-wise Hadamard product. Then the spectral filtering of a graph signal $ \mathbf{x} $ by $ g_\theta $ is 

\begin{equation}
\mathbf{y} = g_{\theta}(\mathcal{L})\mathbf{x}
           = g_{\theta}(\mathbf{U}\Lambda\mathbf{U}^T)\mathbf{x}
           = \mathbf{U}g_{\theta}(\Lambda)\mathbf{U}^T\mathbf{x}.
\end{equation}  

\textbf{Chebyshev approximation for localized filtering.}~~~
The spectral filtering, however, has two limitations: 1) it has high computational complexity of $ \mathcal{O}(n^3) $ due to the eigen-decomposition of the graph Laplacian; 2) it is not localized. Hence, Defferrard et al. propose to use truncated Chebyshev polynomials to approximate the spectral filtering \cite{defferrard2016convolutional}. The K-localized filtering operation is described as follows.

\begin{equation}
\mathbf{y} = g_\theta(\mathcal{L}) \mathbf{x} 
           = \sum_{k=0}^{K-1}\theta_kT_k(\mathcal{L}) \mathbf{x},
\label{eq:chebyshev}
\end{equation}
where $\theta_k$ denotes the $ k $-th Chebyshev coefficient. $T_k(\mathcal{L})$ is the Chebyshev polynomial of order $ k $. It is recurrently calculated by $ T_k(\mathcal{L}) = 2\mathcal{L}T_{k-1}(\mathcal{L}) - T_{k-2}(\mathcal{L})$, where $T_0(\mathcal{L}) = 1,T_1(\mathcal{L}) = \mathcal{L}$. Now the computational complexity is reduced to $ \mathcal{O}(K | \mathcal{E} | ) $.  

\subsection{Feature learning}

Following the graph convolution, we generate a new feature vector for each point from a weight matrix, a bias and the ReLU activation function. This is formulated as follows:

\begin{equation}
    y = \text{ReLU}(g_\theta(\mathcal{L})\vec{x}\mathbf{W} + \vec{b}),
    \label{eq:featureLearning}
\end{equation}
where $\mathbf{W} \in \mathbb{R}^{F_1 \times F_2}$ is the matrix of weight parameters of the trained network, and $ F_1 $ and $ F_2 $ are the dimensions of generated features in two connected layers respectively. $\vec{b} \in \mathbb{R}^{n \times F_2}$ is the bias, while ReLU is an activation function.

In practice, each output feature is calculated by $y_i = \sum\limits_{j=1}^{F_2}w_{i,j}x_i$, $ i=0,...,K-1 $. When $K = 1$, it is equivalent to a one-layer perceptron shared by all the points, which plays an important role in some deep learning networks, such as PointNet. This works well in capturing features of individual points, but loses the neighborhood information. In our model, we take the neighborhood into consideration by graph convolution with truncated Chebyshev polynomials of order $ K > 1 $, thus incorporating local features. 

\begin{figure*}[htbp]
	\centering
    \includegraphics[width=18cm]{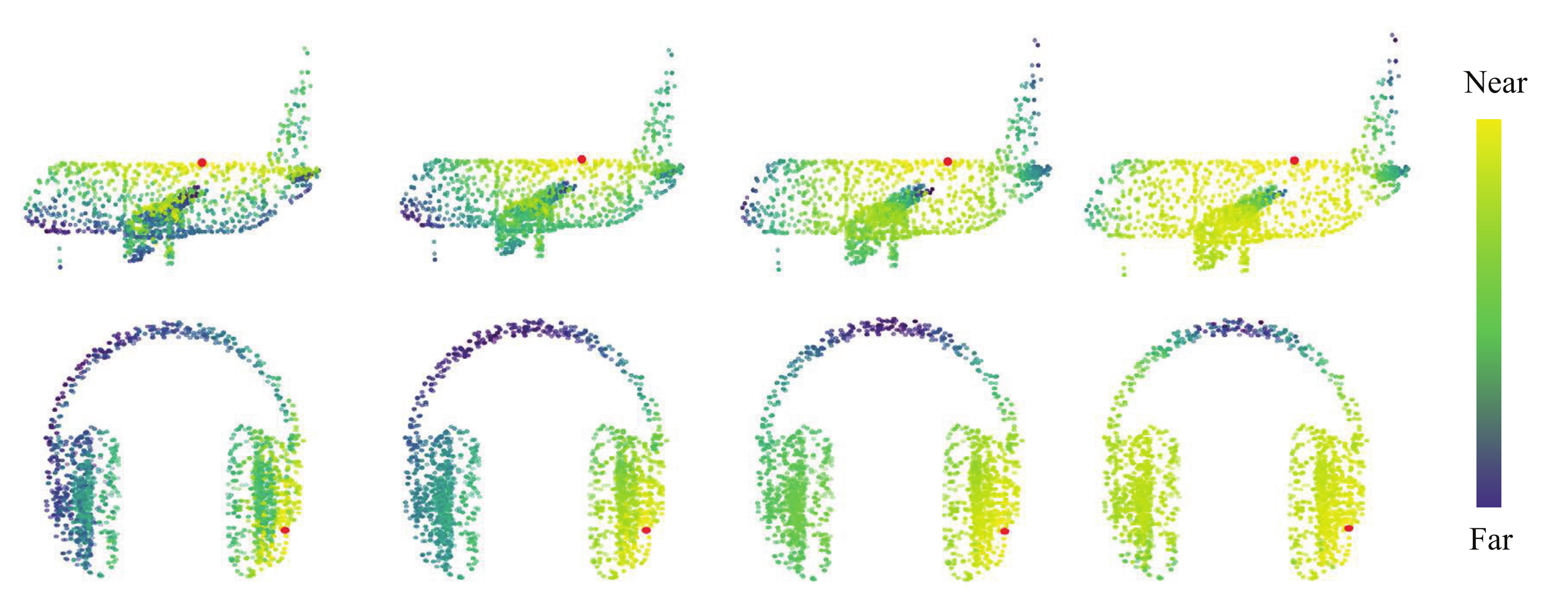}
	\caption{The feature space varies in different layers. The red point in each picture is the basic point. Different colors of the other points represent the Euclidean distance between each point and the basic point in the feature space.}
	\label{fig:dis}
\end{figure*}

Fig.~\ref{fig:dis} demonstrates that the feature space varies in different layers. We observe that a deeper layer is able to capture semantically similar structures better in the high-dimensional feature space.

%% file: theory.tex
This section presents the proposed loss function, in which a graph-signal smoothness prior is added. We then provide theoretical analysis of the added prior, and also prove the permutation invariance property of RGCNN. 

\subsection{The proposed loss function}
While the common error function in the loss function is the consistency of outputs with targets, i.e., the cross entropy, we propose to additionally incorporate a graph-signal smoothness prior as the regularization term. This prior essentially enforces the features of adjacent vertices to be more similar, which eases the segmentation task. 

\textbf{Graph-signal Smoothness Prior.}~~~
A graph signal $ \mathbf{y} $ defined on a graph $ \mathcal{G} $ is smooth with respect to the topology of $ \mathcal{G} $ if
\begin{equation}
	\sum\limits_{i \sim j}a_{i,j}(y_i - y_j)^2 < \epsilon, ~~\forall i,j,
	\label{eq:prior}
\end{equation}
where $ \epsilon $ is a small positive scalar, and $ i \sim j $ denotes two vertices $ i $ and $ j $ are one-hop neighbors in the graph. In order to satisfy Eq.~(\ref{eq:prior}), $ y_i $ and $ y_j $ have to be similar for a large edge weight $ a_{i,j} $, and could be quite different for a small $ a_{i,j} $. Hence, Eq.~(\ref{eq:prior}) enforces $ \mathbf{y} $ to adapt to the topology of $ \mathcal{G} $, which is thus coined \textit{graph-signal smoothness prior}. 

As $ \mathbf{y}^T \mathcal{L} \mathbf{y} = \sum\limits_{i \sim j}a_{i,j}(y_i - y_j)^2 $ \cite{Spielman04}, Eq.~(\ref{eq:prior}) is concisely written as $ \mathbf{y}^T \mathcal{L} \mathbf{y} < \epsilon $ in the sequel. 

\textbf{Loss Function.}~~~
We add the aforementioned graph-signal smoothness prior in the loss function. In particular, the prior is computed from all the three graph convolution layers. The mathematical description is
\begin{equation}
E(\mathbf{y}^o,\mathbf{y}')=-\sum\limits_{i=1}^{n} y_i^o \text{log}({y_i}') 
    + \gamma\sum\limits_{l=0}^{2}\mathbf{y}_l^T\mathcal{L}\mathbf{y}_l,
\end{equation}
where $ \mathbf{y}^o $ is the output score, $ \mathbf{y}' $ is the ground truth label, and $\mathbf{y}_l$ is the feature map of the $ l $-th layer. $\gamma$ is the penalty parameter for the smoothness term, which is empirically set to $ 10^{-9} $ in our experiments.

\subsection{Theoretical analysis}
We provide analysis for the spectral property of the graph-signal smoothness prior and the permutation-invariance property of the proposed architecture. 

\textbf{Theorem 1.}~~~The graph-signal smoothness prior enforces more low-frequency components in the GFT domain. 

\textbf{Proof.}~~~While we have discussed that the added graph-signal smoothness prior regularizes the graph signal to be adapted to the structure of the graph, we further analyze the spectral behaviour of this prior. Specifically, as $ \mathcal{L} $ is diagonalizable as $ \mathbf{U} \Lambda \mathbf{U}^T $ as mentioned earlier, we have

\begin{equation}
\vec{y}^T\mathcal{L}\vec{y} 
= \vec{y}^T (\mathbf{U} \Lambda \mathbf{U}^T)\vec{y}
= (\mathbf{U}^T\vec{y})^T \Lambda (\mathbf{U}^T\vec{y})
= \vec{\alpha}^T \Lambda \vec{\alpha}
= \sum\limits_{i=1}^{n} \lambda_i \alpha_i^2,
\label{eq:theorem1}
\end{equation}
where $ \vec{\alpha} \in \mathbb{R}^n $ is the GFT transform coefficient vector, $ \alpha_i $ is the $ i $-th coefficient, and $ \lambda_i $ is the $ i $-th eigenvalue of $ \mathcal{L} $. The eigenvalues are often sorted as $ \lambda_1 \leq \lambda_2 \leq ... \leq \lambda_n $. 
It is known that larger eigenvalues correspond to higher-frequency GFT coefficients. For instance, $ \lambda_1 $ corresponds to the DC coefficient $ \alpha_1 $, while $ \lambda_n $ corresponds to the highest-frequency coefficient $ \alpha_n $. 

Hence, when we try to minimize the graph-signal smoothness prior in the loss function, the higher-frequency coefficients are weighted by a larger eigenvalue, and are thus penalized more heavily. This means that by adding this prior into the loss function, low-frequency components are better preserved, which leads to smoothing in the spectral domain. The smoothing operation enforces the features of vertices within each connected component of the graph similar, thus greatly easing the segmentation task. 

\textbf{Theorem 2.}~~~The proposed RGCNN is permutation-invariant, i.e., if the rows of the input feature matrix are permuted, the output permutes in the same way. 

Theorem 2 indicates that the segmentation result of RGCNN is irrelevant to the order of the input, which is suitable for the unordered point cloud data. The proof is as follows.

\textbf{Proof.}~~~Denote a $ n \times n $ permutation matrix by $ \mathbf{H} $. We prove if the input $ n \times m $ feature matrix $ \mathbf{P} $ is permuted by $ \mathbf{H} $, i.e., $ \mathbf{H}\mathbf{P} $, then the $ n \times k $ output $ \mathbf{S} $ permutes as $ \mathbf{H}\mathbf{S} $.   

Since the main operation of RGCNN is graph convolution, according to Eq.~(\ref{eq:chebyshev}), we have 

\begin{equation}
\mathbf{S}' = \sum_{k=0}^{K-1}\theta_kT_k(\mathcal{L}) (\mathbf{H}\mathbf{P})
= \mathbf{H} \sum_{k=0}^{K-1}\theta_kT_k(\mathcal{L}) \mathbf{P} 
= \mathbf{H} \mathbf{S}
\end{equation}


Theorem 2 is hence proved.

%% file: results.tex
\begin{table*}[htbp]
    \centering
    \footnotesize
    \caption{Segmentation results on ShapeNet part dataset (in mIoU).}
    \label{table: 1}
    \begin{tabular}{@{}c|c|cccccccccccccccc@{}}
    \toprule
               & mean & aero & bag  & cap  & car  & chair & earphone & guitar & knife & lamp & laptop & motor & mug  & pistol & rocket & skateboard & table \\ \midrule
    ShapeNet   & 81.4 & 81.0 & 78.4 & 77.7 & 75.7 & 87.6  & 61.9     & 92.0   & 85.4  & 82.5 & 95.7   & 70.6  & 91.9 & \textbf{85.9}   & 53.1   & 69.8       & 75.3  \\
    PointNet   & 83.7 & \textbf{83.4} & 78.7 & 82.5 & 74.9 & 89.6  & 73.0     & 91.5   & 85.9  & 80.8 & 95.3   & 65.2  & 93.0 & 81.2   & 57.9   & 72.8       & 80.6  \\
    PointNet++ & \textbf{85.1} & 82.4 & 79.0 & 87.7 & \textbf{77.3} & \textbf{90.8}  & 71.8     & 91.0   & 85.9  & 83.7 & 95.3   & \textbf{71.6}  & 94.1 & 81.3   & 58.7   & 76.4       & \textbf{82.6}  \\ 
    SynSpecCNN & 84.7 & 81.6 & 81.7 & 81.9 & 75.2 & 90.2  & \textbf{74.9}     & \textbf{93.0}   & 86.1  & \textbf{84.7} & 95.6   & 66.7  & 92.7 & 81.6   & \textbf{60.6}   & \textbf{82.9}       & 82.1  \\ \hline
    Ours       & 84.3 & 80.2 & \textbf{82.8} & \textbf{92.6} & 75.3 & 89.2  & 73.7     & 91.3   & \textbf{88.4}    & 83.3 & \textbf{96.0}   & 63.9  & \textbf{95.7} & 60.9   & 44.6   & 72.9       & 80.4  \\ \bottomrule
    \end{tabular}
\end{table*}

In order to evaluate the performance of RGCNN, we carry out extensive experiments for point cloud segmentation, in terms of the segmentation accuracy and the robustness to density and noise. Further, we apply the network architecture to the classification task and provide comparison with the state-of-the-art methods. Finally, we provide analysis on the space and time complexity, and discuss the connections and differences of RGCNN with other competing methods.

\subsection{Experimental setup}

\textbf{Architecture parameters.}~~~ 
In the architecture depicted in Fig.~\ref{fig:graph}, the network comprises of three graph convolution layers with the Chebyshev order $ K = (6,5,3) $ and dimensions of generated features $ F = (128,512,1024) $, followed by three MLP layers $ (512,192,50) $. 


\textbf{Training.}~~~
We conduct experiments on ShapeNet part dataset \cite{yi2016scalable}. This contains 16881 shapes from 16 categories, annotated with 50 labels in total. In the experiments, we first utilize random sampling to extract 2048 points from each model, which form the input point clouds. Then we feed the coordinates and normal of each point into our model as raw features. We follow the training/validation/test setting proposed in \cite{yi2016scalable}, assuming each category label is known for each sample. Our full model is trained on a single Nvidia GeForce GTX 1080Ti with 100 epochs. 

\textbf{Evaluation metric.}~~~
We evaluate segmentation by mean Intersection of Union (mIoU). IoU is widely used in semantic segmentation to measure the ratio of the ground truth and prediction, and mean IoU is the average of IoU for each label appearing in the model categories. We compare our method with ShapeNet~\cite{yi2016scalable}, PointNet~\cite{qi2017pointnet}, PointNet++~\cite{qi2017pointnet++} and SynSpecCNN~\cite{yi2016syncspeccnn}. 

\subsection{Point cloud segmentation results}

\begin{figure*}
    \centering
    \includegraphics[width=\textwidth]{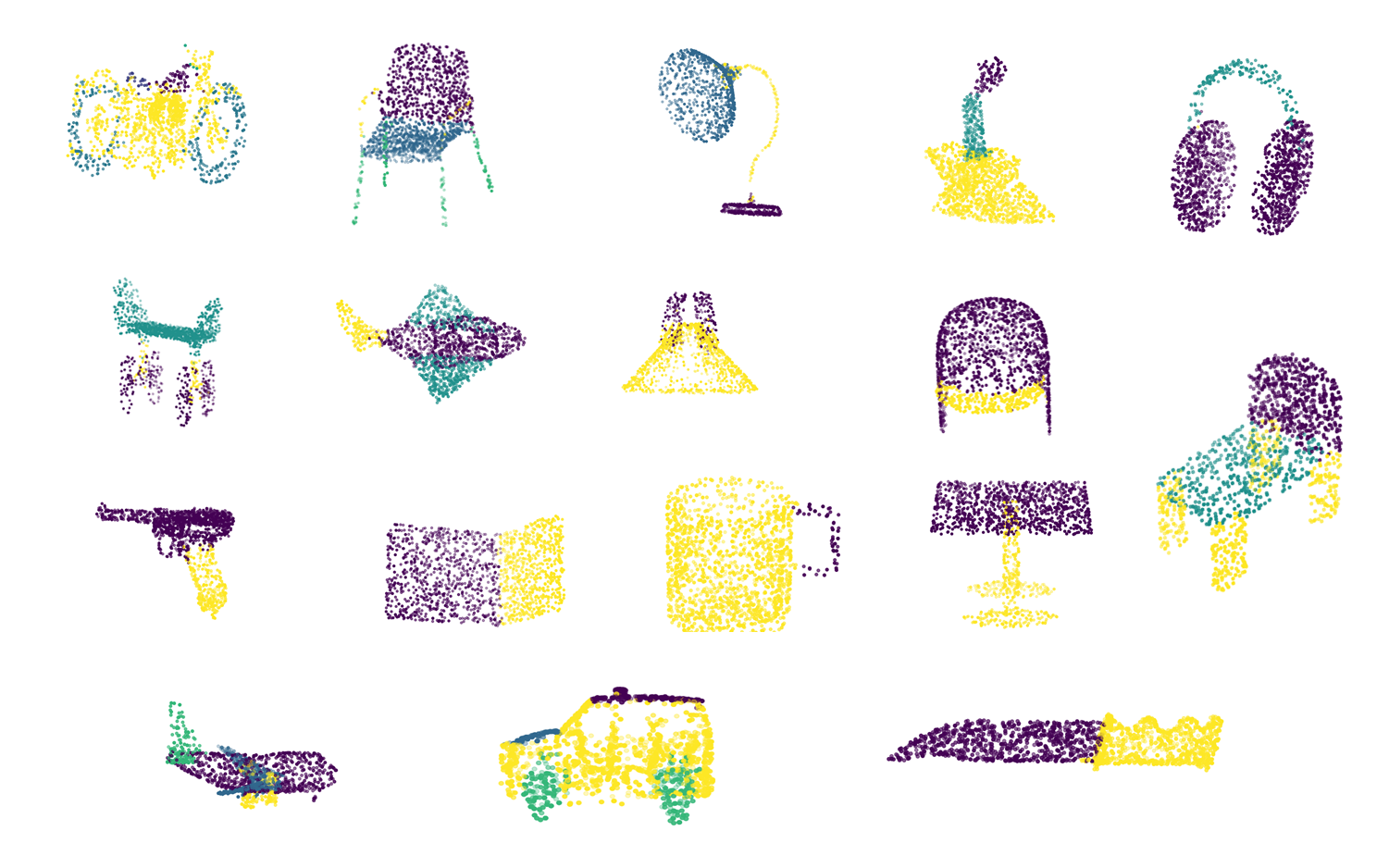}
    \caption{Segmentation results from RGCNN.}
    \label{fig:segResults}
\end{figure*}

The evaluation results are listed in Table~\ref{table: 1}. Our model outperforms the other competing methods in 5 categories, and achieves competitive results with the state-of-the-art. Further, we demonstrate some visual results in Table~\ref{table: visualResults}. It can be observed that RGCNN has better and more consistent segmentation results than PointNet for some challenging objects. More segmentation results of RGCNN are shown in Fig.~\ref{fig:segResults}. 

Also, we evaluate the proposed graph construction of fully-connected graphs. We test the commonly used $k$-nearest-neighbor graphs with $ k=30 $. The resulting mean mIoU is $ 80.4\% $, which is much lower than using the proposed fully-connected graph. This confirms that the fully-connected graph is able to capture more abundant information, thus leading to better segmentation results. 

\begin{table}[]
    \renewcommand\arraystretch{2}
    \centering
    \caption{Visualization of segmentation results.}
    \label{table: visualResults}
    \begin{tabular}{ccc}
    \includegraphics[width=0.1\textwidth]{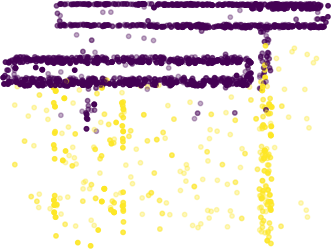} & \includegraphics[width=0.1\textwidth]{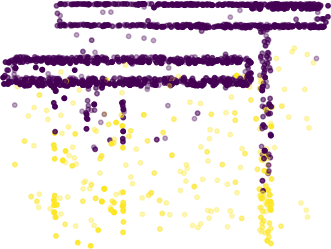} &  \includegraphics[width=0.1\textwidth]{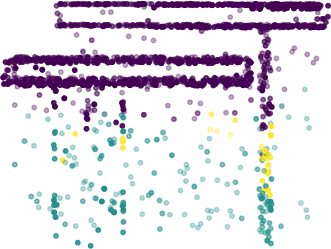}\\
    \includegraphics[width=0.12\textwidth]{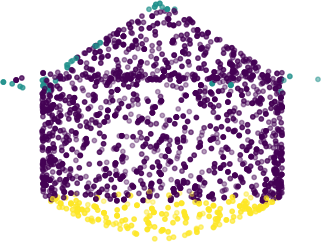} & \includegraphics[width=0.12\textwidth]{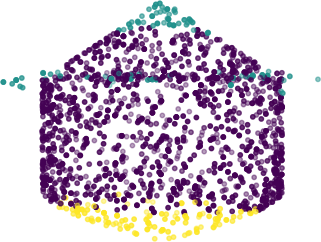} &  \includegraphics[width=0.12\textwidth]{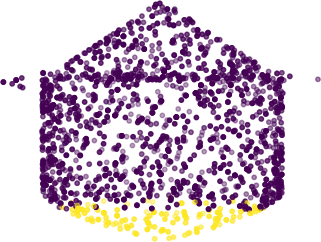}\\
    \includegraphics[width=0.12\textwidth]{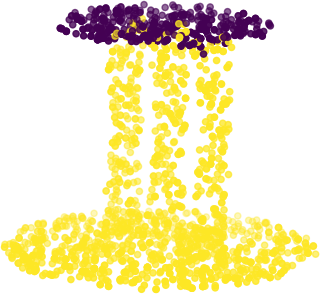} & \includegraphics[width=0.12\textwidth]{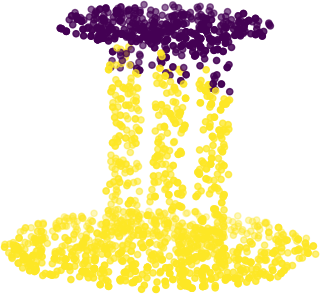} &  \includegraphics[width=0.12\textwidth]{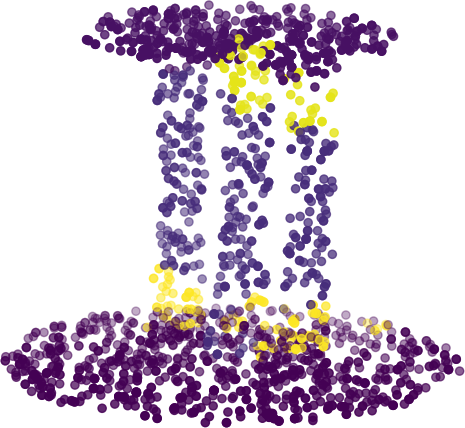}\\
    \includegraphics[width=0.08\textwidth]{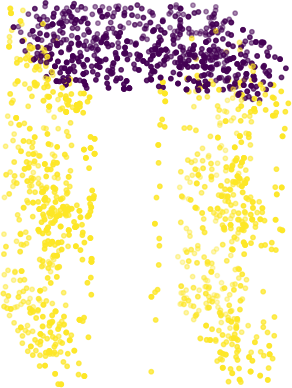} & \includegraphics[width=0.08\textwidth]{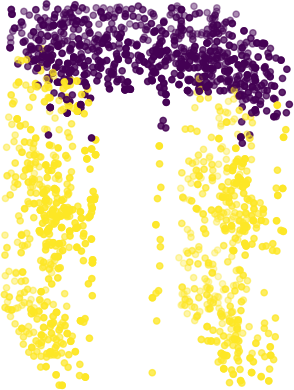} &  \includegraphics[width=0.08\textwidth]{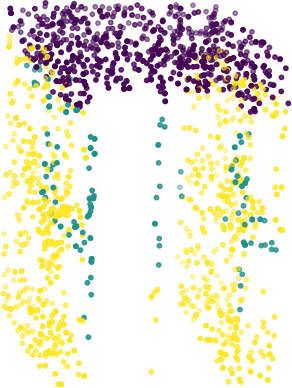}\\
    Ground Truth& Ours & PointNet    \\
    \end{tabular}
\end{table}


\subsection{Robustness test}
\begin{figure}
    	\centering
	    \includegraphics[width=0.35\textwidth]{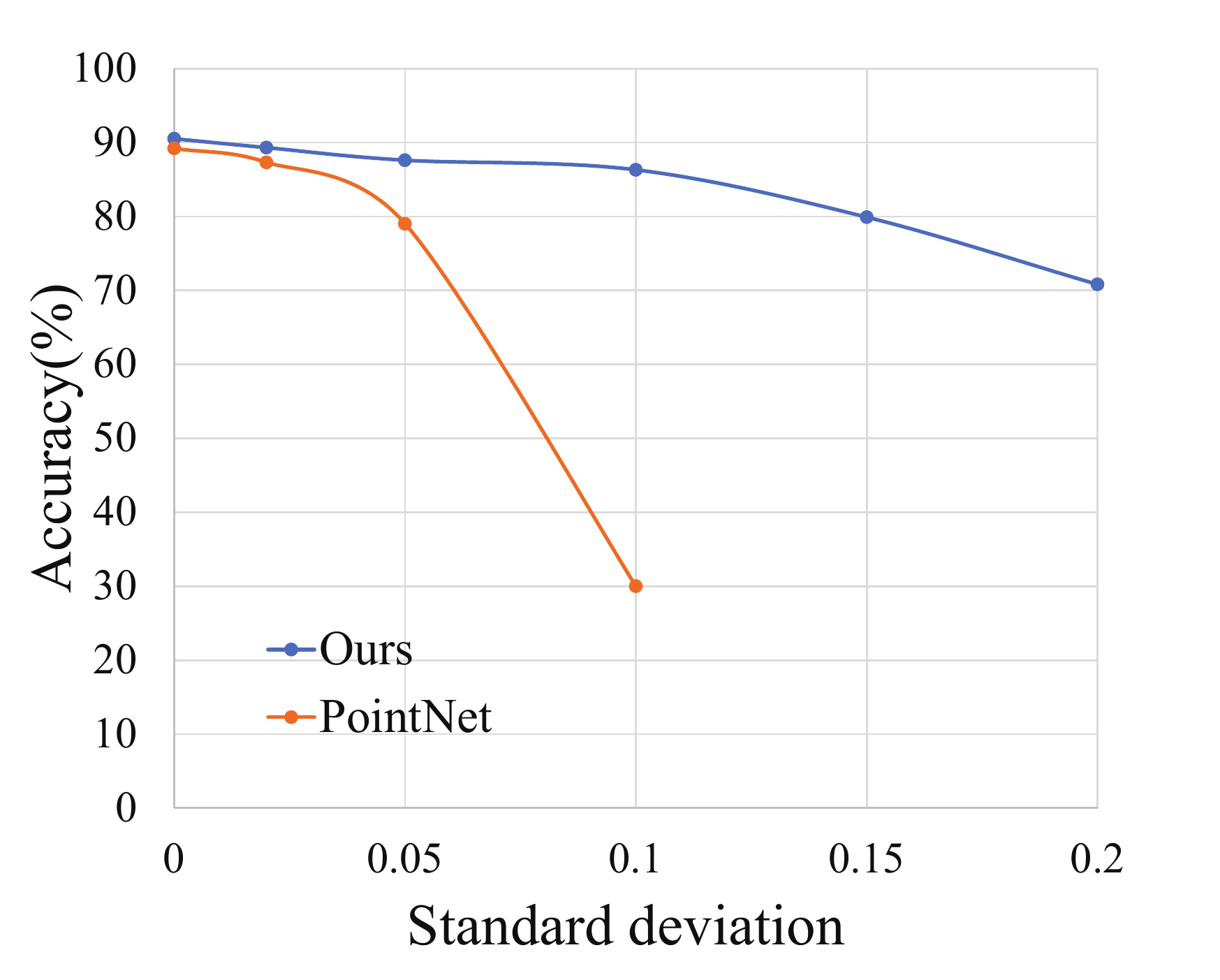}
	    \caption{Accuracy with Gaussian noise.}
	    \label{fig:noise ACC}
\end{figure}

\begin{figure}[htbp]
    \centering
    \subfigure[Ground truth]{\includegraphics[width=0.23\textwidth]{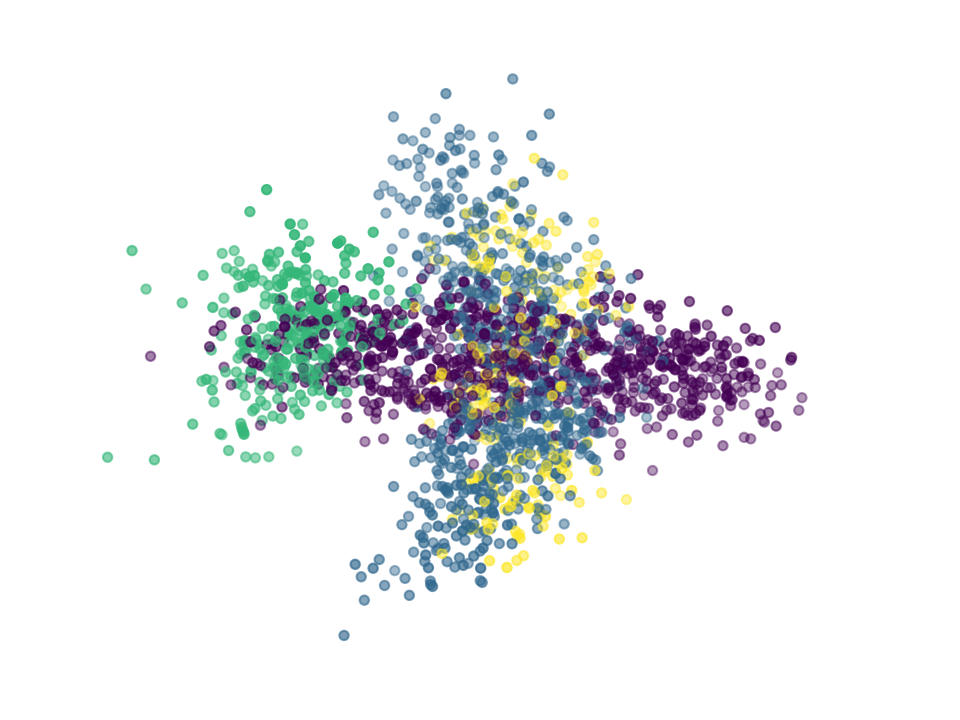}}
	\subfigure[Ours]{\includegraphics[width=0.23\textwidth]{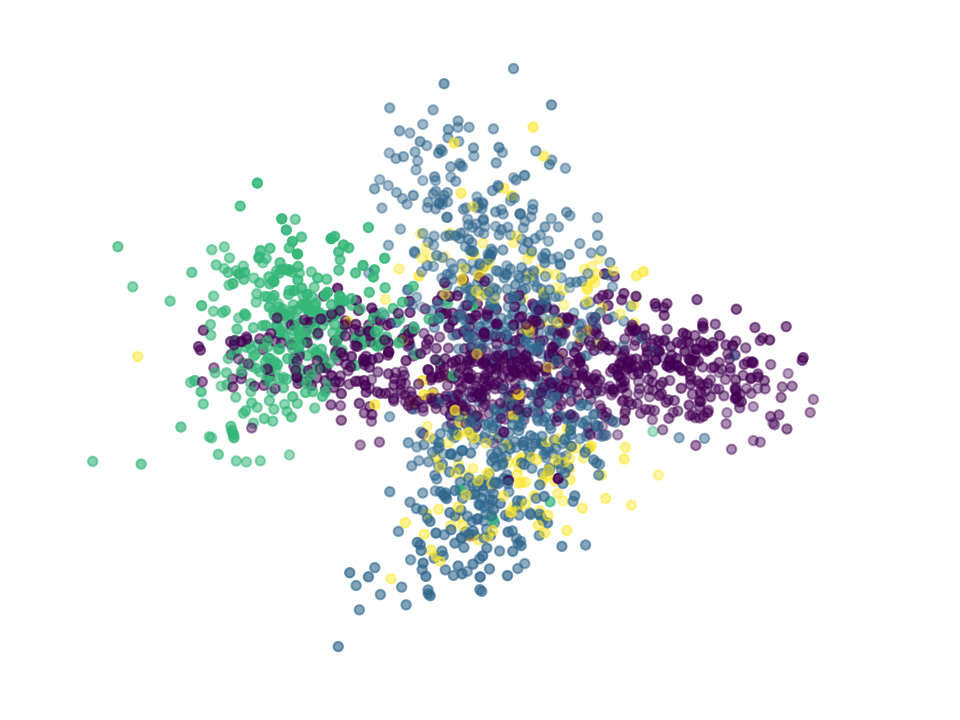}}
	\caption{The comparison between the ground truth and our segmentation result with the input perturbed with the noise level $ \sigma=0.1 $. }
	\label{fig:noise visual}
\end{figure}

\begin{figure}
    \centering
        \centering
        \includegraphics[width=0.35\textwidth]{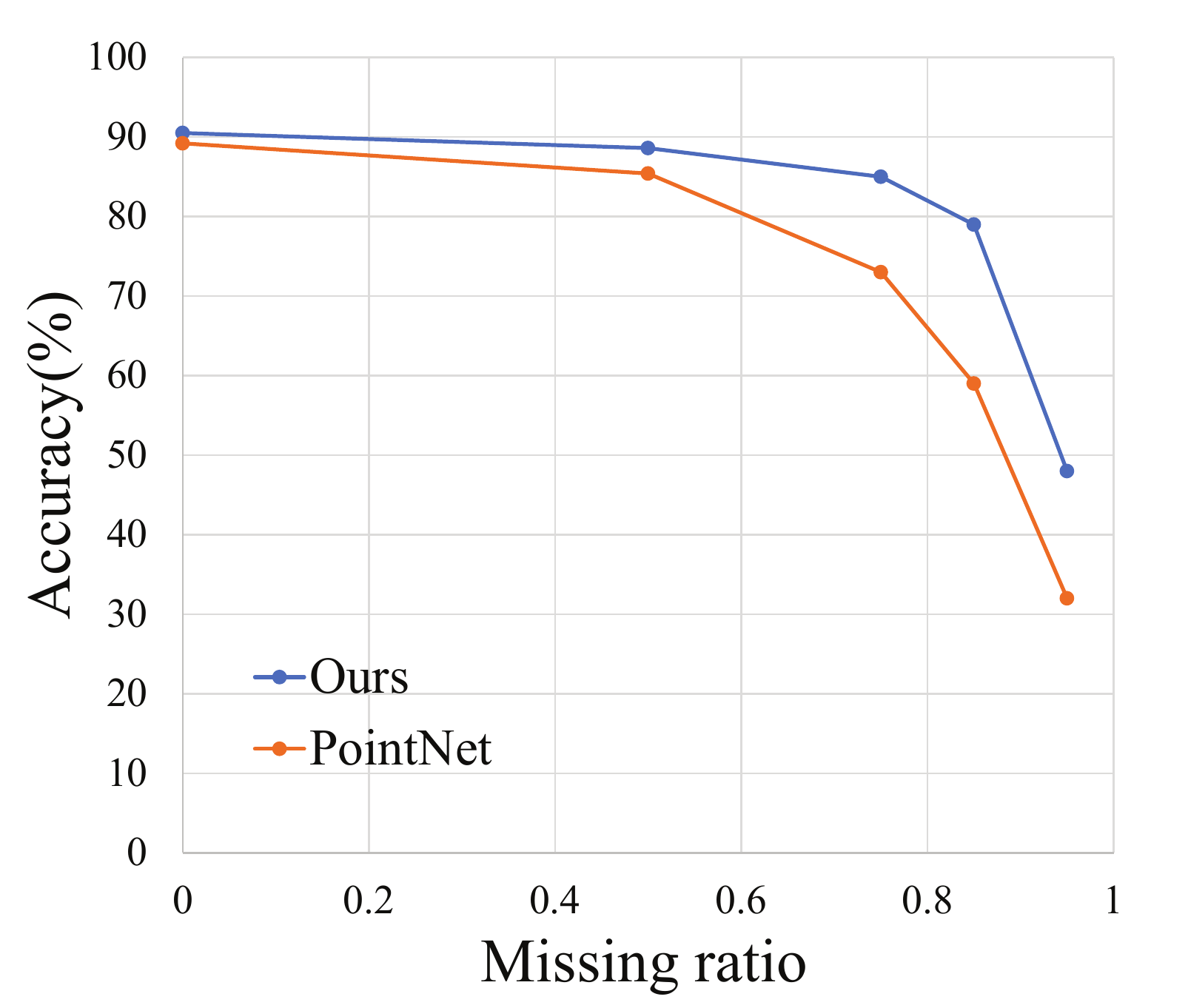}
        \caption{Accuracy with missing data.}
        \label{fig:missing ACC}
\end{figure}

\begin{figure}[htbp]
    \centering
    \subfigure[Ground truth]{\includegraphics[width=0.23\textwidth]{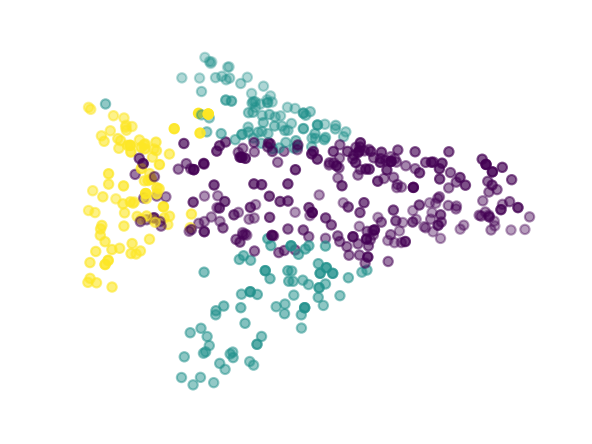}}
	\subfigure[Ours]{\includegraphics[width=0.23\textwidth]{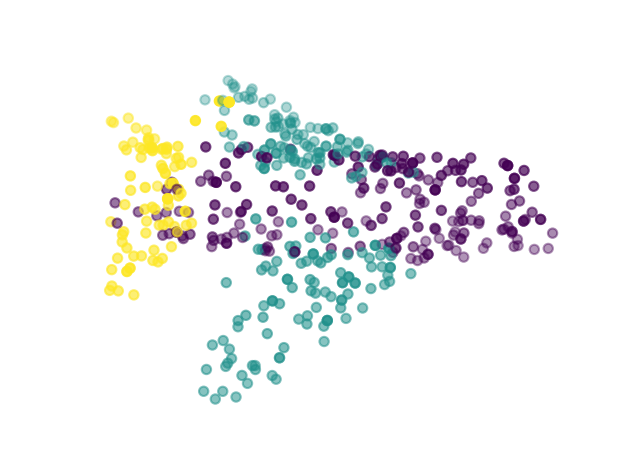}}
	\caption{The comparison between the ground truth and our segmentation result with the input under the missing ratio $ 0.75 $. }
	\label{fig:miss visual}
\end{figure}

\textbf{Robustness to noise}.~~~In order to test the robustness of our model to random noise, we jitter the coordinates of the raw data with Gauss noise, with zero mean and standard deviation $ \sigma \in [0.02,0.2] $. Fig.~\ref{fig:noise ACC} provides mIoU under different noise levels for PointNet and our method. We see that while the performance of PointNet drops quickly with increasing noise variance, RGCNN is robust to noise even when the noise level is high. Also, our segmentation result is visually close to the ground truth from the macroscopic view even when $ \sigma=0.1 $, as demonstrated in Fig.~\ref{fig:noise visual}. 

\textbf{Robustness to density.}~~~We also test the robustness of our model to point clouds of low density. Random dropping is adopted to remove points with missing ratios $ \{0.5,0.75,0.85,0.95\} $. As depicted in Fig.~\ref{fig:missing ACC}, our accuracy keeps $85\%$ even when the missing ratio is $ 0.75 $, which outperforms PointNet ($ 73\% $) significantly. This is also visualized in Fig.~\ref{fig:miss visual}, where our segmentation result is still satisfactory compared with the ground truth.

Hence, RGCNN is very robust to sparse and noisy point clouds. This gives credit to the proposed updated graph Laplacian and graph-signal smoothness prior in the loss function. This property is important in practical applications, since point clouds often suffer from noise or low density mainly due to inherent limitations of acquisition sensors.

\subsection{Application to point cloud classification}
We extend our model to the task of point cloud classification, as shown in the second branch of Fig.~\ref{fig:graph}. It is tested on ModelNet40 dataset to predict the category of a given model. This dataset includes 12311 models from 40 categories, among which we utilize 9843 models for training and 2468 for testing. For each model, we select 1024 points with coordinates and normals randomly as the input point cloud, and then normalize each point cloud to a unit cube. Table~\ref{tb:classification} lists the classification results of different competing methods. It can be seen that our classification accuracy is better than PointNet and comparable to PointNet++.  

\begin{table}[htbp]
    \centering  
    \caption{Classification results.}
    \label{tb:classification}
    \begin{tabular}{ccc}
        \hline
            Metric & Mean Class Accuracy & Overall Accuracy \\
        \hline
            VoxNet\cite{maturana2015voxnet}  & 83.0 & 85.9\\
            PointNet \cite{qi2017pointnet}  & 86.0 & 89.2\\
            PointNet++ \cite{qi2017pointnet++}  & - & 90.7\\
            DGCNN \cite{wang2018dynamic}  & \textbf{90.2} & \textbf{92.2}\\
        \hline
            Ours & 87.3 & 90.5 \\ \bottomrule      
    \end{tabular}
\end{table}

\subsection{Space and time complexity}

We further compare the space and time complexity with other methods. Here, we choose our classification model to test the space and time complexity. Table~\ref{table: 4} shows that our model has the fastest forward time with acceptable model size among these methods. Hence, our model is amenable to real-time classification tasks. Further, we test that the forward time will be even shorter (4.8 ms approximately) if we use fixed graphs instead. 

\begin{table}[htbp]
    \centering
    \caption{Complexity comparison}
    \label{table: 4}
    \begin{centering}
    \begin{tabular}{ccc}
        \hline
        Method & Model Size(MB) & Forward Time(ms)\\
        \hline
        PointNet & 40 & 25.3\\
        PointNet++ & \textbf{12} & 163.2\\
        DGCNN & 21 & 94.6\\
        \hline
        Ours & 22.4 & \textbf{7.5} \\ \bottomrule
        \end{tabular} 
    \end{centering}
\end{table}

\subsection{Discussion}
Finally, we discuss the connections between our method and the other competing methods, as well as the advantages and limit in the following.

\begin{itemize}


\item In graph-based neural networks, the structure of the constructed graph plays an important role for tasks such as point cloud segmentation. Compared with existing graph-based methods in which the graph is fixed in general, our graph structure is dynamic to features in the learning process, thus adaptively capturing the generated features. Also, our graph construction is more computationally efficient. 

\item PointNet deploys MLP to extract the feature of each individual point and utilizes global pooling to extract the global feature, which is a special case in our model when the order of the Chebyshev polynomial is $ K=0 $. Additionaly, our model is able to take the features of the $ K $-hop neighborhood into consideration when $ K \geq 1 $. 

\item In SynSpecCNN, the connection between the spectral and spatial domain is learned, while our graph convolution is another form of spectral approximation but with more flexibility because of the dynamically updated graph Laplacian. 


\item The boundary between two segments is sometimes not quite sharp in our results, which limits the performance to some extent.

\end{itemize}

%% file: conclude.tex
We propose RGCNN, a regularized graph convolutional neural network architecture that directly consumes irregular 3D point clouds. We introduce a graph-signal smoothness prior into the loss function, which essentially enforces Laplacian smoothing in both the spectral and spatial domain. Further, we update the graph Laplacian in each layer of the network in order to adaptively capture the dynamic graphs. Also, we prove the permutation-invariance property of RGCNN, which is suitable for the applications of unordered point clouds. Experimental results on the ShapeNet part dataset for point cloud segmentation validate the effectiveness of RGCNN, showing that RGCNN achieves competitive performance with the state of the art, with much lower computational complexity. We also evaluate that RGCNN is much more robust to both low density and noise than other competing methods. Further, we extend RGCNN for point cloud classification and achieve competitive results on ModelNet40 dataset. 